\documentclass[openacc]{rstransa}

\usepackage[english]{babel}
\usepackage{graphicx}
\usepackage{csquotes}
\usepackage{siunitx}
\usepackage{textcomp, gensymb}
\usepackage{lmodern}

\usepackage{tabularx}
\usepackage[
backend=biber,
style=nature,
sorting=none
]{biblatex}
\titlehead{Research}

\begin{document}

\title{``Golden Ratio Yoshimura'' for Meta-Stable
and Massively Reconfigurable Deployment}

\author{
Vishrut Deshpande$^{1}$, Yogesh Phalak$^{1}$, Ziyang Zhou$^{1}$, Ian Walker$^{2}$, and Suyi Li$^{1}$}

\address{$^{1}$Department of Mechanical Engineering,\\ 
Virginia Tech, Blacksburg, VA USA\\
$^{2}$Department of Electrical Engineering and Computer Science,
University of Wyoming, Laramie, WY USA}

\subject{Themed Issue in Origami and Kirigami Mechanics}


\corres{Yogesh Phalak\\
\email{yphalak@vt.edu}\\
Ziyang Zhou\\
\email{zzhou4@vt.edu}}

\begin{abstract}
Yoshimura origami is a classical folding pattern that has inspired many deployable structure designs. Its applications span from space exploration, kinetic architectures, and soft robots to even everyday household items.  However, despite its wide usage, Yoshimura has been fixated on a set of design constraints to ensure its flat-foldability. Through extensive kinematic analysis and prototype tests, this study presents a new Yoshimura that intentionally defies these constraints. Remarkably, one can impart a unique meta-stability by using the Golden Ratio angle \textcolor{red}{($ \cot^{-1} 1.618 \approx 31.72\degree$)} to define the triangular facets of a generalized-Yoshimura \textcolor{red}{(with $n=3$, where $n$ is the number of rhombi shapes along its cylindrical circumference)}. As a result, when its facets are strategically popped out, a ``Golden Ratio Yoshimura'' boom with $m$ modules can be theoretically reconfigured into $8^m$ geometrically unique and load-bearing shapes. This result not only challenges the existing design norms but also opens up a new avenue to create deployable and versatile structural systems.
\end{abstract}


\begin{fmtext}
\section{Introduction}

Deployable structures have garnered considerable interest over the past decades due to their unique ability to transform from a compact volume to a larger-scale, load-bearing shape \cite{fenci2017, friedman2011a, zhang2021}. Their applications are diverse. For example, they have been the essential components

\end{fmtext}

\maketitle

\noindent for satellites, exoplanet rovers, and space telescopes \cite{puig2010, santiago2013}. They will continue to play a vital role in the future for in-orbit fabrication \cite{liu2022a, ekblaw2019} and space habit construction \cite{yang2023, liu2022b, hill2010}. Here on Earth, deployable structures are ideal for making military and civilian shelters \cite{thrall2014, mira2014} and are even used to develop novel soft robots \cite{troise2021, sedal2020, blumenschein2018, kaufmann2022}. 

A successful deployable structure must meet multiple requirements. It should be efficient in packing (aka. high packing ratio), lightweight for easy transportation, and load-bearing at its final stage. To achieve these goals, researchers have explored various deployment approaches, including pneumatic inflation \cite{friedman2011b, zhang2020, schenk2013}, origami folding \cite{ li2015a, zirbel2014, filipov2015, chen2015, zhai2018}, rolling composite shells \cite{fernandez2017, sickinger2006}, responsive materials \cite{liu2017, wang2017, sun2022}, and multi-stable mechanics \cite{meng2022, bichara2023, haghpanah2016}. Furthermore, one can combine some of these approaches to create innovative solutions. For instance, the combination of pneumatic actuation, bi-stability, and origami folding can enable rapid actuation, akin to the Venus flytrap plant \cite{li2015b}, and allow for meter-scale construction \cite{melancon2021}. Integrating piezo-electric patches onto composite shells can provide versatile actuation and shape control \cite{daye2023}.

However, most deployable structures available today have limited capabilities in that they can only achieve one targeted state. This constraint highlights the potential of a more versatile solution. If a deployable structure can quickly morph into different load-bearing shapes, it can be reconfigured on demand according to the dynamic working environment and desired functionalities. This level of reconfigurability can significantly enhance its appeal. In light of this, we propose and examine a deployable boom structure with massive reconfigurability by expanding the design space of Yoshimura Origami. 

\begin{figure}[b]
    \centering
    \includegraphics[width=0.99\textwidth]{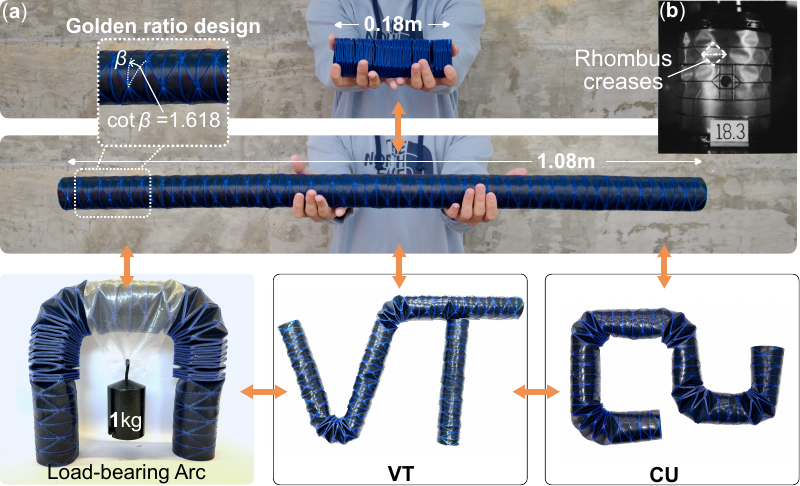}
    \caption{A big-picture overview of the Golden Ratio Yoshimura as a deployable and reconfigurable structure. (a) A 3D-printed, meter-scale prototype can be easily configured into different meta-stable shapes for different purposes, including a fully folded (compact) configuration, a fully deployed configuration, a load-bearing arc, or letters ``VT'' and ``CU.'' Notice the 3D printed sample weighs 225.6 grams, and a second Yoshimura was used to finish the letter ``T.'' (b). The buckling pattern of a cylindrical thin shell reveals the origin of Yoshimura. Image adapted from \cite{singer1982status} with permission.}
    \label{fig:bigpic}
\end{figure}

The Yoshimura pattern is a classical origami design that naturally emerges from the buckling mechanics of thin-wall cylindrical shells \cite{seffen2014} (Fig. \ref{fig:bigpic}). When the cylinder is under axial compression, a uniform tessellation of rhombus-shaped creases can develop on its surface. Each rhombus consists of two triangular facets connected by a fold line along the cylinder's circumference. As the compression continues, the cylinder buckles along these crease lines, and the two triangles in each rhombus fold inward, eventually packing the cylinder into a small volume (Fig. \ref{fig:bigpic}). This buckling-induced crease pattern became the design basis for Yoshimura origami, which has been made into deployable booms \cite{jiang2022, cai2016, pratapa2022, gimenez2024} and soft robotic spines \cite{zhang2021b, yang2024novel, seo2021, xu2021design, zhang2016extensible}.

Despite such wide usage, deployable Yoshimura has been fixated on a design rule in the current literature: The shape of its triangular facets must be constrained by the number of rhombus creases along the cylinder's circumference. This rule can ensure the flat-foldability of Yoshimura. However, we showed that by intentionally breaking these established design rules, one can significantly enrich the kinematic configuration space of Yoshimura and generate a unique ``meta-stability.'' In this way, one can selectively pop out pairs of triangular facets to settle the new Yoshimura into many different elastically stable and load-bearing configurations. \textcolor{red}{Note that this feature sacrifices the flat-foldability, as the more generalized Yoshimura pattern has a finite height when folded.} Remarkably, and \emph{for the first time, we mathematically proved that, for a Yoshimura pattern with three rhombi along its circumference (aka. $n=3$ in Fig. \ref{fig:0pop}), the required triangular facet shape to attain such meta-stability follows the Golden Ratio} (aka. $\cot \beta = \varphi \approx 1.618$ in Fig. \ref{fig:bigpic}, \ref{fig:0pop}). We refer to this new origami design as the ``Golden Ratio Yoshimura,'' and it can become the foundation for those, as mentioned earlier, massively reconfigurable and deployable structures. 

In what follows, section 2 details the design and kinematics of a Golden Ratio Yoshimura module. Section 3 illustrates the meta-stability and configuration space of a larger-scale boom structure with multiple modules. Section 4 presents the fabrication and load-bearing test results of 3D-printed Golden Ratio Yoshimura samples. Finally, Section 5 ends this paper with a conclusion and discussion.

\section{Design and Kinematics of the Golden Ratio Yoshimura \label{sec:Kinematics}}

\begin{figure}[b]
    \centering
    \includegraphics[width=0.9\textwidth]{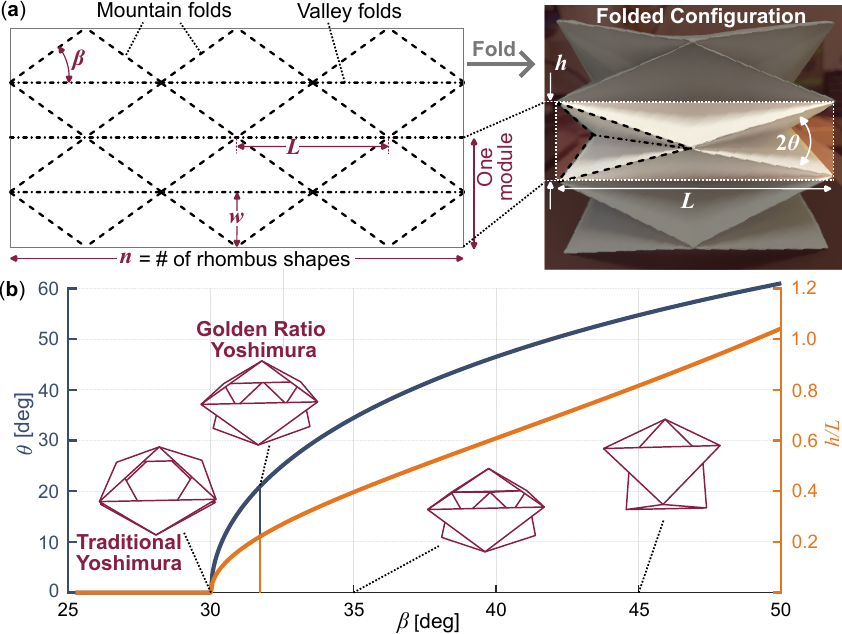}
    \caption{The design and kinematics of folded Yoshimura with a generalized design. (a) The crease pattern design highlights the important parameters. $n$: number of rhombi along the circumference; $L$: length of the valley creases; and $\beta$, sector angle between the mountain and valley creases. (b) The correlations between facets dihedral angle $\theta$, the normalized folded-height $h/L$, and $\beta$.}
    \label{fig:0pop}
\end{figure}

The Yoshimura pattern is a tessellation of \emph{mountain} and \emph{valley} folds such that a 2D sheet, upon folding, curls to form a closed shape resembling a cylinder. Figure \ref{fig:0pop}(a) outlines its layout, where valley creases align with the circumferential direction of the cylinder, and the mountain creases are inclined relative to the valley ones. Such sector angle '$\beta$' defines the Yoshimura pattern's overall shape and will play a central role later in the kinematic analysis. Besides $\beta$, $L$ is the valley crease's length and defines Yoshimura's size; $n$ is the number of rhombi shapes (pairs of triangular facets) along the circumference direction. These three parameters are sufficient to completely define the geometry of the Yoshimura. Moreover, we designate one layer of rhombi creases as a \textit{module}. 

Traditionally, these three design parameters are not independent. Instead, they follow a geometric constraint \cite{zhang2021b}:

\begin{equation}
            \beta = 90\degree - \frac{(n-1)180\degree}{2n} 
            \text{, or }
            2 n \beta = 180\degree.
        \label{eq:unit definition equation}
\end{equation}

For clarity, we call such a design a "traditional Yoshimura." This constraint ensures that, after folding, the Yoshimura is stress-free in its facets and has a zero height (aka flat-foldability). For this study, we focused on $n = 3$ design, so $\beta=30\degree$.

However, \emph{what if Yoshimura defies this constraint}? To this end, this study generalizes the Yoshimura design by setting the sector angle $\beta$ independent of $n$. We discovered that when $\beta$ is larger than the traditional value, the Yoshimura is no longer flat-foldable but has a finite height after folding, denoted as ``folded-height'' $h$ in Fig. \ref{fig:0pop}. Moreover, this height is a function of the dihedral angle $\theta$ between the two triangular facets in each rhombus so that:

\begin{equation}
    h = L \tan\beta \sin\theta.\label{eq:height}
\end{equation}

This dihedral angle in this folded configuration is also related to $\beta$ and $n$ through the following equation:

\begin{equation}
    \tan\beta \cos\theta = \tan\frac{\pi}{2n}. \label{eq:relation_theta_beta}
\end{equation}

The above relationship gives the lower bound for the generalized Yoshimura design: $\beta \geq \frac{\pi}{2n}$, and any smaller $\beta$ value would not be geometrically possible, otherwise $h$ and $\theta$ become negative. The traditional Yoshimura designs precisely follow this lower bound so that $h=0$ and $\theta=0$, giving its flat foldability. On the other hand, when $\beta$ increases, the dihedral angle and folded height increase rapidly (Fig. \ref{fig:0pop}b). For example, when $\beta=45\degree$, $\theta$ reaches 55\degree and $h/L = 0.8$.

\subsection*{2.1 Popping and the Emergence of Golden Ratio Yoshimura \label{sec: 1_2_pop_kinematics}}
\label{sec: kinematics}

The Yoshimura is often used in folded configurations due to its soft and conformable nature for robotic manipulations and locomotion tasks. It is dexterous enough to be manipulated (or actuated) into complex curvatures. However, a fully folded and soft Yoshimura has minimal load-bearing capacity. For the first time, this study introduces ``\textbf{pop-out}'' configurations to the generalized Yoshimura, allowing it to obtain complex and stable shapes with significant load-bearing capacity. The idea is to carefully \emph{buckle-out} the valley creases (or pushing the valley folds in the mid-section of a module outwards) so that the rhombus shape inverts its curvature and stabilizes into a new configuration (Fig. \ref{fig:all-pop}a). Due to this transition, the valley fold in the rhombus ``disappears,'' and a new vertical fold (or bend) emerges. Remarkably, popping out/in a rhombus is an isolated event, and it would not induce popping in the nearby rhombus. Therefore, one can achieve multiple combinations pop-outs and create a \emph{deployable and reconfigurable} structure. In this study focusing on $n=3$, a generalized Yoshimura module has $2^3 = 8$ distinct configurations that can be categorized into four groups: 0 pop-out, 1 pop-out, 2 pop-out, and 3 pop-out. All of these configurations are elastically stable, and we refer to this as ``meta-stability.''  The following sections describe the kinematic characteristics of 1 and 2 pop-out configurations in detail.

\subsubsection*{1 Pop-out Kinematics}
\begin{figure}
    \centering
    \includegraphics[width=0.95\textwidth]{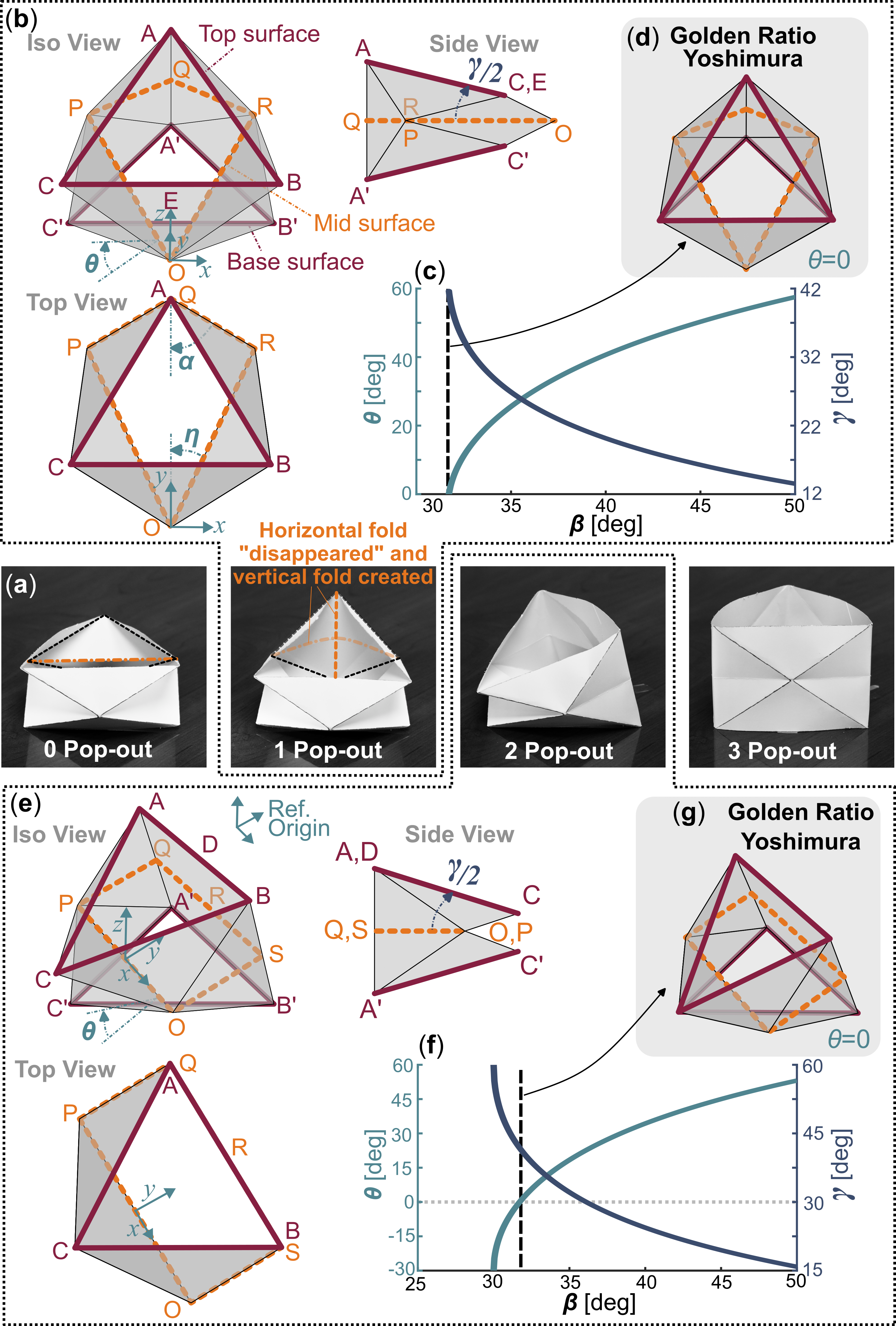}
    \caption{The kinematics of 1 and 2 pop-out Yoshimura modules. (a) Paper models illustrating the sequence of popping out the horizontal valley fold lines to create new configurations.  (b) The detailed view of a 1 pop-out module, highlighting several important geometric features used in the kinematic analysis, such as the top and base interface triangles, mid-surface, tilt angle $\gamma$, and dihedral angle between facets $\theta$. (c, d) The correlation between $\gamma$, $\theta$, and $\beta$, highlighting the Golden Ratio Yoshimura design. (e-g) A similar kinematics analysis on a 2 pop-out module.}
    \label{fig:all-pop}
    \end{figure}

A 1 pop-out configuration emerges when a single valley crease is pushed outwards from the mid-section of a module. Figure \ref{fig:all-pop}(b) details the geometry of a 1 pop-out module from three different viewing angles. The isometric view highlights three essential features to explain its kinematics. A top interface triangle ABC, a mid-surface OPQR, and a base interface triangle A'B'C'. Here, \emph{we made a critical assumption that the shape of the top and base interface triangle remain fixed and equilateral, regardless of the popping configurations}. In this way, one can seamlessly assemble multiple modules at different configurations into a more complex structure (which we will discuss in the later sections.)
    
The top view in Figure \ref{fig:all-pop}(b) clearly shows the kite-shaped mid-surface OPQR. Here, the valley crease PR is popped out, and we needed two angles to define this kite shape: $\alpha$ and $\eta$, which follows the relationship: 

\begin{equation}
    \sin\eta = \frac{\sin\alpha}{2}.
    \label{eq:relation eta,alpha}
\end{equation}  

\textcolor{red}{Here, $\eta$ must lie in $[0,\frac{\pi}{3}]$ to ensure a real-valued solution for angle $\alpha$.} To further correlate these two angles to the dihedral facet angle ($\theta$) and mountain crease incline angle $\beta$, we started by calculating the height of the original triangular facet ($w$ as in Fig. \ref{fig:0pop}), in that:

\begin{equation}
    w = \frac{L}{2} \tan \beta.
    \label{eq:half-height definition eq}
\end{equation}

Then, using vertex O as the reference point, we can calculate the positions of vertices A, B, and R relative to O:

\begin{equation}
    \begin{split}
        O &\equiv \left[0,0,0\right], \quad(\text{Reference}) \\
        A &\equiv \left[0, L\cos\eta + \frac{L}{2}\cos\alpha, w\right], \\
        B &\equiv \left[\frac{L}{2},w\cos\theta, w\sin\theta\right], \\
        R &\equiv \left[L\sin\eta, L \cos\eta, 0\right].
    \end{split}
\end{equation}

Based on the above-mentioned assumption on fixed interface triangles, the distance between vertices A and B does not change after pop-out; therefore

\begin{equation}
        |A-B|^2 = L^2 =
        \left(\frac{L}{2} - 0\right)^2 + \left( w \cos\theta - \left(L\cos\eta + \frac{L}{2}\cos\alpha\right)\right)^2 + \left(w \sin\theta - w \right)^2.
\end{equation}

Simplifying this equation with the help of Eq. (\ref{eq:half-height definition eq}) yields:

\begin{equation}
    2 \tan^2\beta(1-\sin\theta) + \left(2\cos\eta + \cos\alpha\right)^2 - 2\tan\beta \cos\theta(2\cos\eta + \cos\alpha) = 3.
    \label{eq:dist1}
\end{equation}

Similarly, the distance between the vertices R and B does not change so that,

\begin{equation}
        |R-B|^2 = \left(\frac{L}{2 \cos\beta}\right)^2 =
        \left(L \sin\eta - \frac{L}{2}\right)^2 + \left(L\cos\eta - w \cos\theta \right)^2 + \left(0 - w\sin\theta \right)^2.
\end{equation}

Simplifying again with the help from Eq. (\ref{eq:half-height definition eq}) yields,

\begin{equation}
    \tan\beta \cos\theta = \frac{1 - \sin\eta}{\cos \eta}.
    \label{eq:dist2}
\end{equation}

Using Eq. (\ref{eq:relation eta,alpha}-\ref{eq:dist2}), we could fully define the geometry of a 1 pop-out Yoshimura module. These equations are highly non-linear and transcendental, so they were solved numerically. Figure \ref{fig:all-pop}(c) summarizes the numerical solution showing the correlation between the dihedral facet angle $\theta$ and the Yoshimura design parameter $\beta$. \textcolor{red}{It is worth noting that, in the 0 pop-out configuration as shown in Fig. \ref{fig:0pop}, the dihedral angle $\theta$ is the same for all three pairs of triangular facets. However, in the 1 pop-out configuration, $\theta$ becomes $90\degree$ for the popped facets (aka. PAR and PA'R pair in Fig. \ref{fig:all-pop}(b)), while the dihedral angle $\theta$ of the other two facet pairs (OBR-OB'R and OCP-OC'P) are non-zero and related to $\beta$}. 

\textcolor{red}{To understand the lower limit of sector angle $\beta$ that allows this kinematically admissible configuration, we obtain the closed-form solution of $\beta$ when the dihedral angle $\theta$ vanishes. By substituting zero dihedral angle condition ($\theta=0$), we observe from the top view of Fig. \ref{fig:all-pop}(b) that $\eta = \frac{\pi}{2} - 2\beta$. Combining this observation with Eq. (\ref{eq:relation eta,alpha}) implies that $\beta$ is constrained to the interval $\left[\frac{\pi}{12}, \frac{\pi}{4}\right]$. Applying these conditions to Eq. (\ref{eq:dist2}) and using the identity $\cos\left(\frac{\pi}{2} - x\right) = \sin(x)$ yields:}

\begin{equation}
    2 \tan^2\beta + \left(2\sin(2\beta) + \cos\alpha\right)^2 - 2\tan\beta \left(2\sin(2\beta) + \cos\alpha\right) = 3. \label{eq:simp_pop1}
\end{equation}

\textcolor{red}{
We then substitute $\alpha$ from Eq. (\ref{eq:relation eta,alpha}) into Eq. (\ref{eq:simp_pop1}), leading to a polynomial in $\tan\beta$:}

\begin{equation}
    \left(\tan \beta -\varphi^{-1}\right)\left(\tan \beta -\varphi\right)\left( 5\tan^4\beta + 4\sqrt{5}\tan^3\beta - 2\tan^2\beta - 4\sqrt{5}\tan\beta - 7 \right) = 0. \label{eq:1pop_pol}
\end{equation}

\textcolor{red}{
Here, $\varphi$ denotes the Golden Ratio, given by $\frac{1 + \sqrt{5}}{2}$. Given the constraints on $\beta$, $\tan\beta$ must lie within the interval $[2 - \sqrt{3}, 1]$. Consequently, the only valid solution for Equation (\ref{eq:1pop_pol}) is $\tan\beta = \frac{1}{\varphi}$. This implies $\cot\beta = \varphi$, which leads to $\beta \approx 31.72^\circ$. Hence, no solution exists for $\beta < 31.72^\circ$. This result is intrinsically linked to the Golden Ratio and guided us to define the "Golden Ratio Yoshimura" pattern, where the dihedral angle $\theta$ of the folded facet pairs reduces to zero.} As a result, the vertices B and C on the top interface triangle will coincide with their counterpart B' and C' on the base triangle (i.e., the top surface and base triangle share a common edge). For any $\beta < 31.72\degree$,  the two adjacent facets PCO and PC'O would interfere and create geometric frustration, resulting in unreliable mechanics.

The tilt angle ($\gamma$) is defined as the angle between the top and base interface triangle.  It can be calculated with the help of a vector joining the mid-point of BC and vertex A. The relative coordinates of the mid-point of BC (let's say E) as,

\begin{equation}
    E \equiv \left[0,w\cos\theta, w\;\sin\theta\right],
\end{equation}
    
Now, the angle between this vector EA and the mid-section plane is half of the tilt angle, which can be found as,

\begin{equation}
    \tan \frac{\gamma}{2} = \frac{w - w\sin\theta}{L\cos\eta + \frac{L}{2} \cos\alpha - w\cos\theta} \textcolor{red}{= \frac{1 - \sin{\theta}}{2\tan{\beta}\cos{\eta} + \tan{\beta} \cos\alpha - \cos\theta}}. \label{eq:tilt_1_pop_out}
\end{equation}

The above equation explains the relation of the tilt angle for a generalized Yoshimura that depends on the pattern angle, $\beta$, dihedral angle, $\theta$, and the angle $\eta$. This tilt angle for our Golden-ratio Yoshimura is $(\gamma_\text{gold})_{1-p/o} \approx 41.8\degree$, and it is interestingly the maximum for the $n=3$ design.

\subsubsection*{2 Pop-out Kinematics}

A 2 pop-out configuration emerges when two valley creases in the mid-section of a module are pushed outwards. One can reach this configuration either by popping out an additional valley fold from a 1 pop-out state or popping in a rhombus from the 3 pop-out configuration. Fig. \ref{fig:all-pop}(e) details the geometry of this configuration with three different views. Again, we assume the top and base interface triangles remained unchanged. 

From the top view, one can see that mid-surface OPQSR---where the valley creases PR and RO are popped out --- has a square shape, which is relatively more straightforward to calculate than the 1 pop-out case. To correlate the dihedral facet angle $\theta$ with the facet sector angle $\beta$, we start by calculating the positions of vertices B and C relative to the mid-point of PO crease so that:

\begin{equation}
    \begin{split}
        C &\equiv \left[0,-w \cos\theta, w\sin\theta\right], \\
        B &\equiv \left[\frac{L}{2}, \frac{L}{2}, w\right].
    \end{split}
\end{equation}

Here, the distance between the vertices B and C is conserved according to the fixed interface triangle assumption,

\begin{equation}
    |B-C|^2 = L^2 = \left(\frac{L}{2} - 0\right)^2 + \left(\frac{L}{2} - (-w \cos\theta)\right)^2 + \left(w - w\sin\theta\right)^2.
\end{equation}
    
Simplifying this equation yields the following nonlinear relationship,

\begin{equation}
    \tan^2 \beta (1 - \sin\theta) + \tan \beta \cos\theta = 1. \label{eq:2pop}
\end{equation}

\textcolor{red}{The dihedral angle ($\theta$) between facet PCO and PC'O can drop to zero since any negative $\theta$ will cause self-intersection between the two facets and are not physically possible. Substituting $\theta=0$ into Eq. (\ref{eq:2pop}) yields a quadratic equation of $\tan{\beta}$. Upon solving it with a quadratic formula and choosing the solution with a positive value, we again obtain $\tan{\beta} = \frac{-1+\sqrt{5}}{2} = \varphi^{-1}$, or $\beta\approx31.72^\circ$.}


\emph{Therefore, we \textcolor{red}{obtain the closed-form analytical solution} that the Golden Ratio angle is the lower bound of the $\beta$ value for kinematically admissible pop-out operations} ($n=3$).

To calculate tilt angle $\gamma$, we consider a vector from the mid-point of crease AB to vertex C. Its angle with the mid-surface is half of the $\gamma$ angle. The coordinates of this mid-point (let's say D) relative to the mid-point of crease OP are,

\begin{equation}
    D \equiv \left[0,\frac{L}{2},w\right]
\end{equation}
    
Hence, the tilt angle is defined by,

\begin{equation}
    \tan\frac{\gamma}{2} = \frac{w-w\sin\theta}{\frac{L}{2}+w\cos\theta} \textcolor{red}{= \frac{1-\sin{\theta}}{\tan{\beta}+\cos{\theta}}}. \label{eq:tilt_2_pop_out}
\end{equation}

The correlation between $\gamma$ and $\beta$ is presented in Fig. \ref{fig:all-pop}(f). Interestingly, the maximum tilt angle also occurs at the Golden Ratio design, and it is the same as that for the 1 pop-out configuration, i.e., $(\gamma_\text{gold})_{2-p/o} \approx 41.8\degree$. Although the orientation of this tilt angle is different from the 1 pop-out configurations (more on this in the following section).

\section{Configuration Space of A Golden Ratio Yoshimura Boom \label{sec:ConfigSpace}}

\subsection*{3.1. Forward Kinematics Formulation}

\begin{figure}[b]
    \centering
    \includegraphics[width=0.95\textwidth]{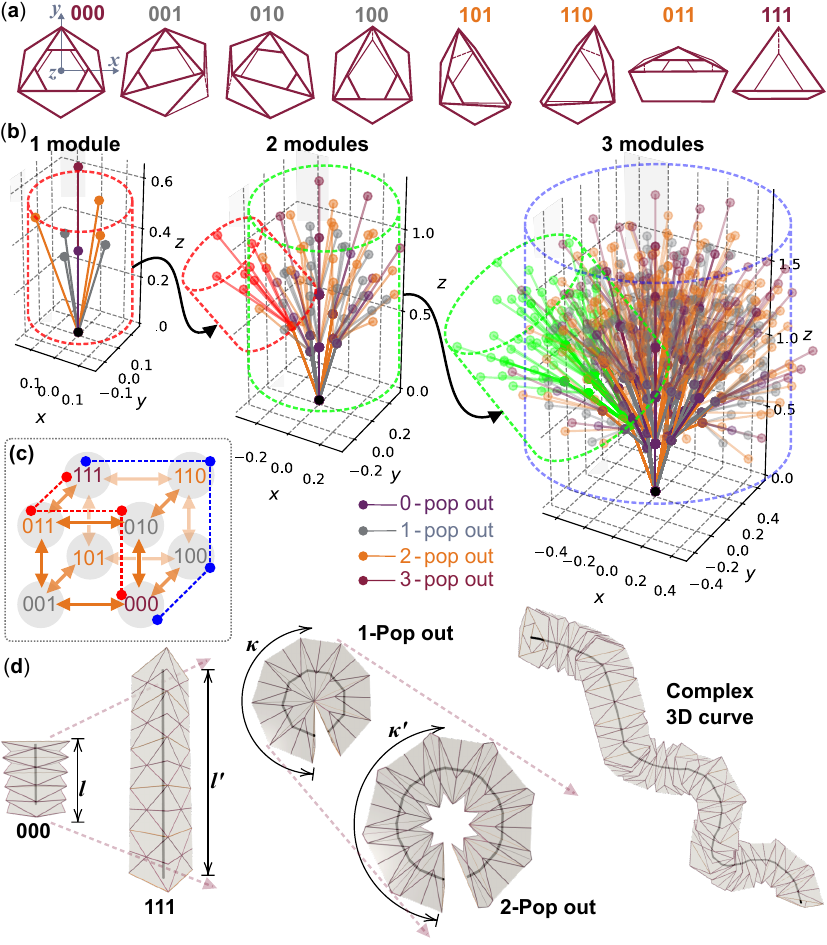}
    \caption{Kinematics and configuration space of the multi-modular Golden Ratio Yoshimura boom: (a) Illustration of the eight distinct meta-stable states in the binary `pop state' representation. (b) Plots comparing the achievable configuration from Yoshimura boom with 1, 2, and 3 modules, showing a fractal nature. (c) Representation of the state transition scheme using a 3-bit Gray code. (d) Examples of length scaling (left), curvature scaling (center), and a complex curve (right) that are achievable with Yoshimura booms.}
    \label{fig:config-space}
\end{figure}

The previous section's kinematic analysis on the Golden Ratio Yoshimura shows its modules' state depends on the number of \textit{popped-out} facets and the relative orientation of the adjacent modules. Here, the equilateral and fixed triangular interface shared by two adjacent modules is an essential feature of this arrangement, allowing for seamless and independent assembly of all possible states. We used a binary ``pop state'' representation, where ``1'' corresponds to a pop-out (deployed) facet, and ``0'' represents a pop-in (folded) facet. For the $n=3$, there are a total of $2^3=8$ distinct states, as shown in Fig. \ref{fig:config-space}(a).

Naturally, the following fundamental question arises: \textit{How can we derive a comprehensive model to analyze the kinematic configuration space of a large-scale Golden Ratio Yoshimura boom with multiple modules?} To this end, we formulated a \textit{modified Denavit-Hartenberg (mDH)} convention that is typically used in robot kinematics \cite{craig2006introduction}, essentially treating each module like a robotic link. First, we defined a coordinate frame $\mathbf{r}_i$ attached to module \#$i$'s base interface triangle, where the vertex A' of the base triangle lies on the $y-$axis, and the side formed by the remaining vertices is parallel to the $x-$axis (highlighted in Fig. \ref{fig:all-pop}g). All dimensions are re-scaled so that the side length of the interface triangle equals 1 unit. 

To calculate the position and orientation of the next coordinate frame $\mathbf{r}_{i+1}$ --- which defines the top interface triangle of module \#$i$ or the base interface triangle of the next module \#$i+1$ (Fig. \ref{fig:all-pop}g) --- one must apply two consecutive transformations based on Yoshimura's kinematics model: One from the base triangle A'B'C' to the mid-surface ($^\text{base}\mathbf{T}_\text{mid}$), and the other from the mid-surface to the top interface triangle ABC ($^\text{mid}\mathbf{T}_\text{top}$). 

First, assuming the Yoshimura module's configuration is symmetric about the $y-z$ reference plane, which applies to states 000, 100, 011, and 111 in Fig. \ref{fig:config-space}(a), one can write this transformation matrices as:

\begin{equation} \label{eq:T1}
    ^{i}\mathbf{T}_{i+1} =  ^\text{base}\mathbf{T}_\text{mid} \cdot ^\text{mid}\mathbf{T}_\text{top},
\end{equation}

\noindent where 

\begin{align}
    ^\text{base}\mathbf{T}_\text{mid} &= \mathbf{Rot}_x\left(\gamma/2\right) \mathbf{Tr}_x(0) \mathbf{Rot}_z(0) \mathbf{Tr}_z\left(d/2\right), \\
    ^\text{mid}\mathbf{T}_\text{top} &= \left[ \mathbf{Rot}_x\left(-\gamma/2\right) \mathbf{Tr}_x(0) \mathbf{Rot}_z(0) \mathbf{Tr}_z\left(-d/2\right)\right]^{-1}.
\end{align}

Here, $\mathbf{Rot}_k$ and $\mathbf{Tr}_k$ are the rotation and translation matrices along the $k-$coordinate axes ($k=x, y$ or $z$).  $\gamma$ is the tilt angle between the top and base interface triangles, as defined in Fig. \ref{fig:all-pop}, and $d$ is the distance between the center points of these two triangles (also referred to as the ``slant height''). For example,

\begin{equation}
    \mathbf{Tr}_x(s_x) = \begin{bmatrix}
            1& 0& 0& s_x \\
            0& 1& 0& 0 \\
            0& 0& 1& 0 \\
            0& 0& 0& 1 \\
        \end{bmatrix},
\end{equation}

\begin{equation}
    \mathbf{Tr}_z(s_z) = \begin{bmatrix}
        1& 0& 0& 0 \\
        0& 1& 0& 0 \\
        0& 0& 1& s_z \\
        0& 0& 0& 1 \\
    \end{bmatrix},
\end{equation}

\begin{equation}
    \mathbf{Rot}_x(\phi_x) = \begin{bmatrix}
        1& 0& 0& 0\\
        0& \cos{\phi_x}& -\sin{\phi_x}& 0\\
        0& \sin{\phi_x}& \cos{\phi_x}& 0\\
        0& 0& 0& 1\\
    \end{bmatrix},
    \end{equation}
    
\begin{equation}
    \mathbf{Rot}_z(\phi_z) = \begin{bmatrix}
        \cos{\phi_z}& -\sin{\phi_z}& 0& 0\\
        \sin{\phi_z}& \cos{\phi_z}& 0& 0\\
        0& 0& 1& 0\\
        0& 0& 0& 1\\
    \end{bmatrix}.
\end{equation}

Here, $s_x$, $s_z$, $\phi_x$, and $\phi_z$ are generic length and angle variables. If the Yoshimura module's configuration is not symmetric about the $y-z$ reference plane (which applies to meta-stable states 001, 010, 101, and 110 in Fig. \ref{fig:config-space}a), one can first rotate the module about its $z-$axis at the base by a ``phase angle'' $\psi$ until becoming symmetric, apply the transformation defined in Eq. (\ref{eq:T1}), and then rotate back.  Therefore, a more generic transformation matrix becomes:

\begin{equation}
^{i+1}\mathbf{T}_{i} = \mathbf{Rot}_z(\psi) \mathbf{Rot}_x\left(\gamma/2\right) \mathbf{Tr}_z(d) \mathbf{Rot}_x\left(\gamma/2\right) \mathbf{Rot}_z(-\psi).   
\end{equation}

More specifically,

\begin{equation}
    ^{i+1}\mathbf{T}_{i} =  \begin{bmatrix}   
    \cos^2{\psi} + \cos{\gamma}\sin^2\psi & \frac{1}{2}(1-\cos{\gamma})\sin{2\psi} & \sin{\gamma}\sin{\psi} & d\sin{\frac{\gamma}{2}}\sin{\psi}\\
    \frac{1}{2}(1-\cos{\gamma})\sin{2\psi} & \cos{\gamma}\cos^2{\psi} + \sin^2{\psi} & -\sin{\gamma}\cos{\psi} & -d\sin{\frac{\gamma}{2}}\cos{\psi}\\
    -\sin{\gamma}\sin{\psi} & \sin{\gamma}\cos{\psi} & \cos{\gamma} & d\cos{\frac{\gamma}{2}}\\
    0 & 0 & 0 & 1
    \end{bmatrix}.
\end{equation}

The corresponding transformation parameters for each module are listed in Table \ref{table: Kinematical Parameters} \textcolor{red}{(Detailed calculations of parameters are given in \ref{Apndx:FK_params})}. Finally, by treating each module as an independent building block, we could determine the kinematic state of the entire structure, by applying a chain of transformations from the $i^{th}$ and $j^{th}$ modules ($i < j$):

\begin{equation}\label{eq:T2}
    \mathbf{x}_j = ^j\mathbf{T}_{j-1} {}^{j-1}\mathbf{T}_{j-2} \dots ^{i+1}\mathbf{T}_{i} \cdot \mathbf{x}_i,
\end{equation}

\noindent whereas $\mathbf{x}_j$ and $\mathbf{x}_i$ are the vertices coordinates of $j^{th}$ and $i^{th}$ module.

\begin{table}[b]
\centering
\renewcommand{\arraystretch}{2.3}
\setlength{\extrarowheight}{-3pt}
\caption{Forward kinematics model parameters for each meta-stable state. Note that $\varphi$ represents the golden ratio, i.e. $\varphi= \frac{1+\sqrt{5}}{2} \approx 1.618$. All angles are given in radians.}
\label{table: Kinematical Parameters}
\begin{tabular}{c|c c c}
\hline
Pop in/out state &$\psi$ (phase angle) &$\gamma$ (tilt angle) &$d$ (slant height)\\
\hline
\hline
$000$ &$0$ &$0$ &$\frac{1}{\varphi}\sqrt{1-\frac{\varphi^2}{3}} $\\ \hline
$001$ &$-\frac{2\pi}{3}$ &$2 \sin^{-1}\left(\frac{1}{\sqrt{3}\varphi}\right)$ &$\frac{1}{3\varphi}$\\ \hline
$010$ &$\frac{2\pi}{3}$ &$2 \sin^{-1}\left(\frac{1}{\sqrt{3}\varphi}\right)$ &$\frac{1}{3\varphi}$\\ \hline
$011$ &$0$ &$-2 \sin^{-1}\left(\frac{1}{\sqrt{3}\varphi}\right)$ &$\frac{2}{3\varphi}$\\ \hline
$100$ &$0$ &$2 \sin^{-1}\left(\frac{1}{\sqrt{3}\varphi}\right)$ &$\frac{1}{3\varphi}$\\ \hline
$101$ &$\frac{2\pi}{3}$ &$-2 \sin^{-1}\left(\frac{1}{\sqrt{3}\varphi}\right)$ &$\frac{2}{3\varphi}$\\ \hline
$110$ &$-\frac{2\pi}{3}$ &$-2 \sin^{-1}\left(\frac{1}{\sqrt{3}\varphi}\right)$ &$\frac{2}{3\varphi}$\\ \hline
$111$ &$0$ &$0$ &$\frac{1}{\varphi}$\\
\hline
\end{tabular}
\end{table}

\subsection*{3.2. Exploring and Accessing the Configuration Space of Yoshimura}

Since the modules are independent of each other as they switch between meta-stable states, the achievable configuration space of a deployable Yoshimura can grow exponentially as its number of modules increases. Using the transformation Eq. (\ref{eq:T2}), we can calculate the reachable workspace of a single Yoshimura module by calculating the positions of its top interface triangle's center at the eight meta-stable states (the first plot of Fig. \ref{fig:all-pop}b). With each additional module, the workspace 'replicates' and 'grows,' similar to a fractal tree canopy. This self-similarity allows for diverse shapes and configurations, as illustrated in the second and third plots in Fig. \ref{fig:all-pop}(b). As a result, even with a few modules, a substantial portion of space becomes accessible, emphasizing the Golden Ratio Yoshimura's massive deployability and reconfigurability.

Adding more modules to the Yoshimura boom presents a challenge in actuation; attaching one actuator to every facet might create an overly complicated mechatronic setup. To solve this problem, we suggest using a single actuator that can travel freely through a hollow cavity inside the Yoshimura and along its spline. This way, the actuator can selectively pop out or in individual facets, one at a time. While developing such an actuator is out of the scope of this paper, we formulated a \textit{3-bit Gray code} to establish an efficient state transition scheme for a single module, as demonstrated in Fig. \ref{fig:config-space}(c). This code ensures that adjacent states differ by only one switch, providing a single face-popping action per step. As a result, following this Gray code minimizes the actuation requirements. \textcolor{red}{For example, in the lattice visualization of 3-bit Gray code, there are two minimum actuation paths to switch from the state 000 to 111, one shown in blue and one in red}. One could repeat the same scheme iteratively when dealing with structures containing multiple modules (e.g., using a \textit{6-bit Gray code} for two modules and a \textit{3$n$-bit Gray code} length for $n$ modules). 

The Golden Ratio Yoshimura boom offers a versatile way to construct complex shapes by combining different modules. We can adjust the length scaling and curvature to achieve intricate shapes overall. For example, we could reach the maximum increment in length ($l'$) by stacking either the 111 module or a pair of 1 and 2 pop-out modules with the same phase $\psi$ (e.g., 001 and 110 according to Table \ref{table: Kinematical Parameters}). The corresponding increment in length is $1/\varphi \approx 0.618$. On the other hand, the minimum increment in length ($l$) is achieved with 000 modules, and the value is given by $1/\varphi \sqrt{1 - \varphi^2/3} \approx 0.2205$.

Curvature also varies across modules. Yoshimura's curvature can be calculated as the ratio of tilt angle $\gamma$ to slant height $d$. Based on the kinematics model and its results summarized in Table \ref{table: Kinematical Parameters}), Modules 000 and 111 exhibit zero curvature. 2 pop-out modules have the intermediate curvature, and 1 pop-out module has maximum curvature. Stacking different models with curvatures in the same plane could achieve varying $\kappa$ in that,

\begin{equation}
    \kappa = \frac{\text{total angle swept}}{\text{average arc-length}} = \frac{\sum_i{\gamma_i}}{\sum_i{d_i}}.
\end{equation}

\textcolor{red}{In Fig. \ref{fig:config-space}(d)  the various combinations of module configurations are depicted. These simulations are made by triangulating nodes calculated using the kinematic model and then plotted using a 3D visualizer based on Python.}

\section{Experimental Assessments}
It's worth noting that besides the rich configuration space via meta-stability, another significant advantage of the Golden Ratio Yoshimura boom is that each meta-stable state should be load-carrying (except for the fully folded 0 pop-out one). This is because the popped-out facets take up the shape of a cylindrical shell. Evidently, a more robust fabrication method is necessary to assess such load-carrying performance. Although the paper-folded samples used in the previous sections exhibit kinematic properties similar to those predicted by the theoretical model, they are too fragile to carry any meaningful payloads. To this end, we developed a customized fabrication technique using dual-material 3D printing. 

\subsection*{4.1. Sample Fabrication via 3D Printing}

\begin{figure}[b]
    \centering
    \includegraphics[scale=0.95]{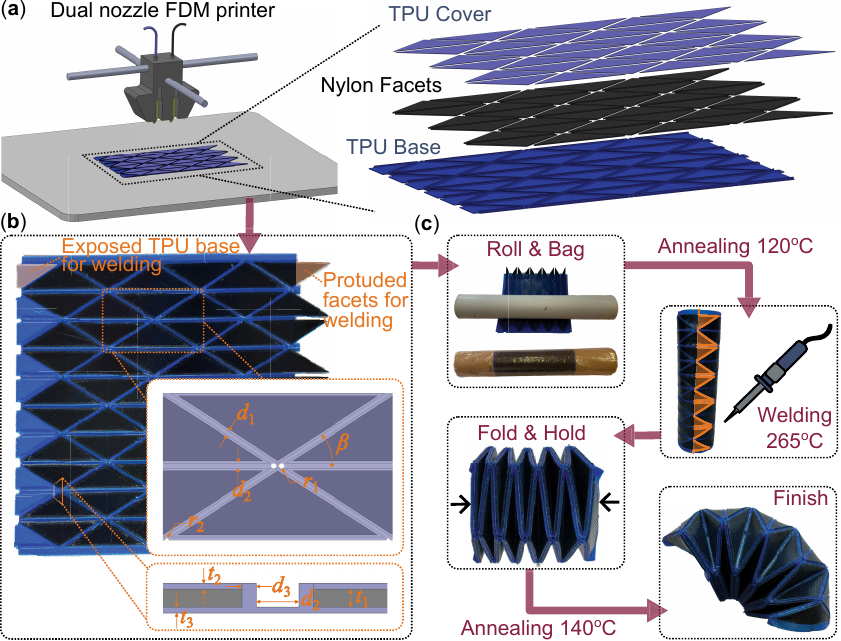}
    \caption{3D printing and fabrication process of a Golden Ratio Yoshimura sample. (a) Diagram of a flat Yoshimura sample being printed using the layer-by-layer approach. (b) Detailed design parameter of the 3D-printed sample. (c) Fabrication steps after printing.  }
    \label{fig:fab}
\end{figure}

In this study, we employed the widely-used Fused Deposition Modeling (FDM) 3D-printing technique to fabricate the Golden Ratio Yoshimura samples (dual-material Ultimaker S5 printer). As outlined in {Fig. \ref{fig:fab}, the Yoshimura structure is printed flat first and then folded, reflecting the origami paper folding philosophy. However, unlike paper, the 3D-printed model utilizes two materials: stiff nylon and soft thermoplastic polyurethane (TPU) 95A. Here, nylon is used for the triangular facets, significantly enhancing Yoshimura's meta-stability and overall stiffness. TPU 95A primarily functions as the crease lines, ensuring sufficient flexibility for folding and reconfiguration. It is important to note that creating a strong bond between these two materials during the printing process poses significant challenges, and our repeated trials and errors showed that nylon has the best compatibility with TPU 95A. To further strengthen the bonding, we took inspiration from prior works \cite{xue2024rigid} and developed a layer-by-layer printing method. First, we reduced the 3D printing's layer thickness to 0.1mm. Then, we printed the flat Yoshimura pattern in three parts: a TPU base, nylon facets, and a TPU cover (Fig. \ref{fig:fab}a,b). The TPU base --- with a thickness ($t_3$) 0.4mm and a zig-zag printing path --- is a durable foundation for the entire structure. Thin TPU walls ($d_{3}$) built on top of this base create space for the nylon facets and provide lateral connections between nylon and TPU. The nylon facets are 0.6mm thick ($t_{1}$). Finally, a 0.2mm thin TPU cover ($t_{2}$) securely seals the nylon facets within the structure, ensuring proper bonding even under significant load and deformation.

Overall, the 3D-printed sample used thicker and stiffer nylon to define the facets and relied on the thinner and softer TPU 95 to define the fold lines. Its design aims to closely replicate the kinematics model (e.g., the printed facet sector angle matched the Golden Ratio $\beta=31.7\degree$). Still, it differs from the paper-folded sample, which has a negligible thickness and sharp folds. So, we made a few design accommodations. For example, we designed a gap, labeled as ($d_1$ and $d_{2}$), between the thick facets to facilitate proper folding. We also strategically placed holes of radius $r_{1}$ or $r_2$ at the vertices to reduce stress concentration.  Table \ref{table: print} summarizes the design parameters selected for 3D printing.

Once printing was complete, the Yoshimura sample was wrapped around a plastic tube and covered with a plastic film for annealing. Note that the tube's radius was large enough so that the Yoshimura sample would not overlap itself, and the plastic film cover was tight. Then, the sample was annealed in a temperature-controlled oven at 120\degree C for 5 minutes (EasyComposites OV301). This step created a curvature in the Yoshimura to facilitate the subsequent heat welding (Fig. \ref{fig:fab}c). We used a manual soldering gun set at 265\degree C to weld the curved Yoshimura sample into a closed cylinder. It is critical to point out that the triangular nylon facets on one side of the Yoshimura sample protrude out of the base TPU layer, and this protrusion directly matches the exposed TPU base layers on the other side (highlighted in Fig. \ref{fig:fab}b,c). This design maximized the surface area for heat welding while ensuring the finished specimen's overall thickness and mechanical properties are consistent between welded and unwelded portions.

Once the welding was complete, we folded the Yoshimura sample along the crease lines into the fully folded 0 pop-out state. We custom-built a frame to compress and hold the sample and then put everything into the oven for another annealing at 140\degree C for 15 minutes. This was a crucial step in reducing tension along the crease lines to facilitate folding. After cooling to room temperature, the sample remained fixed to the frame for an additional 30 minutes to eliminate any residual stress. Following these procedures, the sample was removed from the frame, and as shown in Fig. \ref{fig:fab}(c), the Yoshimura sample exhibited meta-stability and retained its configuration once the desired facets were popped out.

\begin{table}
\begin{center}
\caption{Design Parameters for 3D printed Yoshimura sample}\label{table: print}
\begin{tabularx}{0.75\textwidth}{>{\centering\arraybackslash}X|>{\centering\arraybackslash}X>{\centering\arraybackslash}X}
 \hline
 \textbf{Parameter} & \textbf{Value} \\
 \hline\hline
 Triangular facet angle $\beta$ & 31.72$^\circ$   \\
 \hline
 Gap width for folding $d_{1}$ & 1.0mm  \\ 
 \hline
 Gap width for folding $d_{2}$ & 1.5mm  \\
 \hline
 Vertical TPU wall thickness $d_{3}$ & 0.5mm  \\
  \hline
 Stress relieving hole radius $r_{1}$ & 1.5mm   \\ 
 \hline
 Stress relieving hole radius $r_{2}$ & 1.0mm  \\
  \hline
 Nylon facet thickness $t_{1}$ & 0.6mm   \\ 
 \hline
 Top TPU layer thickness $t_{2}$ & 0.2mm  \\
 \hline
 Base TPU layer thickness $t_{3}$ & 0.4mm  \\
 \hline
\end{tabularx}

\end{center}
\end{table}

\subsection*{4.2 Load Bearing Capacity Tests}
We conducted longitudinal compression and 3-point bending tests on the 3D-printed Yoshimura samples to assess their mechanical load-bearing capacity.

\begin{figure}[b] 
    \centering
    \includegraphics[scale=0.9]{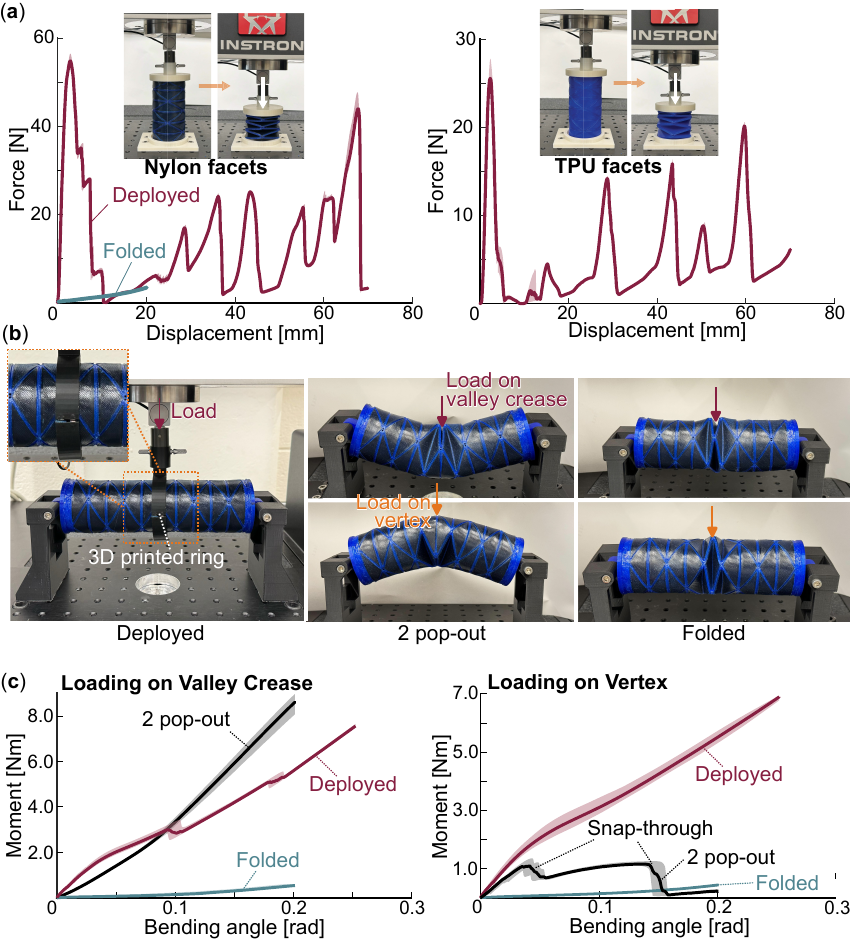}
    \caption{Experiment Testing on 3D-printed Golden Ratio Yoshimura's load-bearing capacity. (a) Longitudinal compression test results, highlighting the differences between a dual-material sample and a pure TPU sample. (b) Details of the experimental setups for the 3-point bending tests. (c) 3-point bending test results of the 3D printed Yoshimura sample, which was bent either via the valley crease or via the vertex. Here, the solid lines are averaged results, and the accompanying shaded regions are standard deviations of 7 consecutive loading cycles.}
    \label{fig:exp}
\end{figure}

\subsubsection*{Longitudinal Compression Test}

Two different Yoshimura samples were fabricated for the compression tests: The first sample had a dual-material setup with nylon facets, while the second was entirely printed from TPU 95. The two samples shared identical geometry with three modules. Comparing the load-bearing capabilities of these two samples could help us investigate how the material composition influences the overall structural performance. 

\textbf{Fig. \ref{fig:exp}}(a) summarizes the longitudinal compression test results (Instron 6500 with 100N load cell). Overall, the 3D-printed Yoshimura samples \textcolor{red}{showed robust performance}, exhibiting repeatable mechanical behaviors even after extensive usage. The dual-material Yoshimura sample withstood over 50 Newtons of compression force before beginning to buckle and transitioning from the deployed 3 pop-out state to the folded 0 pop-out state eventually (the sample itself weighs about 23 grams). Such buckling and folding occurred in sequence, so each small peak on the force-deformation curve indicates an inward buckling (or pop-in) of one triangular facet pair. The Yoshimura sample fully folded after being compressed by 62mm, and once it transitioned into the folded state, the Yoshimura sample behaved like a soft linear spring. In contrast, the pure TPU Yoshimura sample could carry only about 25 Newtons of compression force. These compression tests demonstrated that the nylon facets can significantly enhance the overall load-bearing, so we only used the dual material sample for the subsequent 3-point bending tests. 

\subsubsection*{3-Point Bending Test}
For the 3-point bending test, we designed a Yoshimura sample with five modules. The central module was the focus of the test: It was switched between 0 pop-out, 2 pop-out, and 3 pop-out states, while the four modules at both ends remained in the fully pop-out configuration. This setup ensured consistent boundary conditions when the input force was applied to the central module. Moreover, we bent the Yoshimura sample in two opposite directions: one involving loading on a valley crease and the other on a vertex. A 3D-printed ring fixture distributed the transverse load more evenly over Yoshimura's surface (Fig. \ref{fig:exp}b).

Fig. \ref{fig:exp}(c) summarizes the tested external moment-bending angle relationships, elucidating the bending stiffness of the Yoshimura structure under various configurations. Overall, the Yoshimura could carry significantly more bending load in their 2 pop-out and 3 pop-out configurations than the 0 pop-out state, validating its advantage as a load-carrying reconfigurable structure. The bending response is generally linear, except for the 2 pop-out results. When loaded on the valley crease, the 2 pop-out Yoshimura exhibits a close-to-linear bending response. However, when loaded on the vertex, the 2 pop-out Yoshimura showed a significantly nonlinear behavior with two distant ``snap-through'' events. These snap-through events occurred when the popped-out facets, under compression in this loading condition, eventually buckled and folded inwards. This behavior is similar to what was observed in the longitudinal compression test. 

\section{Conlusion and Discusion}
By intentionally breaking the traditional design rules of Yoshimura origami, this study presents a new ``Golden Ratio Yoshimura'' as the backbone for new deployable and massively re-configurable structures. The conventional Yoshimura design requires the sector angle of its triangular facets ($\beta$) to be directly related to the number of rhombus patterns ($n$) along its circumference. However, we generalized the design in this study by setting $\beta$ as an independent variable. For the first time, we proved mathematically that when $n=3$, the more general Yoshimura pattern would obtain meta-stability when the facet sector angle is larger than the Golden Ratio: $\beta \geq 31.7\degree=\cot^{-1} 1.618$. With such meta-stability, one can selectively pop out pairs of triangular facets of the Yoshimura and transition it between many elastically stable states. On the other hand, if the $\beta<31.7\degree$, these meta-stable states become kinematically inadmissible.

Each pop-out/in operation is independent, meaning that an elementary module of the $n=3$ Golden Ratio Yoshimura has $2^3=8$ meta-stable states. As a result, a deployable Yoshimura Boom structure with $m$ modules can theoretically achieve $8^m$ geometrically unique and load-bearing shapes. Furthermore, we demonstrated the practical use of Golden Ratio Yoshimura by fabricating meter-scale prototypes using multi-material FDM 3D printing. These prototypes, with their soft TPU 95A base and stiff Nylon facets, validated the potential of the Golden Ratio Yoshimura boom to be re-configured into a wide variety of shapes, each suited for different tasks. 

Philosophically speaking, this study's results highlight that even a seemingly minor tweak to the decade-old origami design rules could unveil significant new potential, potentially fostering a new generation of deployable and re-configurable structures.

\appendix   

\section{Forward Kinematics Parameters \label{Apndx:FK_params}}

\subsection*{Phase Angles (\texorpdfstring{$\psi$}{Lg})}

\textcolor{red}{As explained in Section \ref{sec:ConfigSpace}, configurations 000, 100, 011, and 111 are symmetric about the y-z reference plane; hence the phase angle is 0. On the other hand, the asymmetry of 001, 010, 101, and 110 is compensated by a rotation about the z-axis with an angle of $2\pi/n$. In this study, $n=3$, so that the phase angle is $2\pi/3$. Based on Fig. \ref{fig:config-space}(a), one can see that the 010 and 101 configurations need a rotation by $2\pi/3$ to become symmetric about the y-z reference plane. Similarly, 001 and 100 configurations need a rotation of $4\pi/3$ or $-2\pi/3$.}

\subsection*{Tilt Angles (\texorpdfstring{$\gamma$}{Lg})}

\textcolor{red}{The tilt angles for the 000 and 111 configurations are 0 as the top and bottom planes are parallel. Furthermore, both the 1 pop-out configurations (001, 010, 100) and the 2 pop-out configurations (101, 110, 011) are congruent in terms of tilt, resulting in equal magnitudes of tilt angles in these cases. The magnitudes are calculated using Eq. (\ref{eq:tilt_1_pop_out}) for 1 pop-out and Eq. (\ref{eq:tilt_2_pop_out}) for 2 pop-out.}

\textcolor{red}{For the 1 pop-out configuration, one can substitute $\theta = 0$, $\beta = \cot^{-1}\varphi$, and the values of $\alpha$ and $\eta$ calculated in Section \ref{sec: 1_2_pop_kinematics} ($\eta = \pi/2 - 2\beta$, $\alpha = \tan^{-1}2$) to simplifying Eq. (\ref{eq:tilt_1_pop_out}), yielding:}

\begin{equation}
    \textcolor{red}{\tan\left(\frac{\gamma}{2}\right) = \frac{\varphi}{1 + \varphi},}  
\end{equation}

\noindent \textcolor{red}{which further simplifies based on the trigonometric identity $\sin{x} = \frac{\tan x}{\sqrt{1 + \tan^2 x}}$:}

\begin{equation}
    \textcolor{red}{|(\gamma_\text{gold})_{1-p/o}| = 2 \sin^{-1}\left(\frac{1}{\sqrt{3}\varphi}\right) \approx 41.81^\circ.}   
\end{equation}

\textcolor{red}{Interestingly, substituting $\theta = 0$ into Eq. (\ref{eq:tilt_2_pop_out}) yields the exact same solution:}

\begin{equation}
    \textcolor{red}{|(\gamma_\text{gold})_{2-p/o}| = 2 \sin^{-1}\left(\frac{1}{\sqrt{3}\varphi}\right) \approx 41.81.}
\end{equation}

\textcolor{red}{Since the symmetric configurations 100 and 011 are tilted opposite to each other, we assign positive values to the 1 pop-out configuration and negative values to the 2 pop-out configuration.}

\subsection*{Slant Heights ($d$)}

\textcolor{red}{For the 0 and 3 pop-out configurations, the slant height is the same as the module's total height.}

\textcolor{red}{The height at the 000 configurations can be calculated using the equation for $h$ (Eq. (\ref{eq:height})) and substituting the $\theta$ solution from Eq. (\ref{eq:relation_theta_beta}) corresponding to the Golden Ratio, i.e., substituting $\beta = \cot^{-1}\varphi$, so that:}

\begin{equation}
    \textcolor{red}{(d_\text{gold})_{0-p/o} = \frac{1}{\varphi} \sin\left(\cos^{-1}\left(\varphi \tan\left(\frac{\pi}{6}\right)\right)\right) =     \frac{1}{\varphi} \sqrt{1 - \frac{\varphi^2}{3}} \approx 0.2205.}
\end{equation}

\textcolor{red}{For the 1 and 2-pop out configurations, the slant height is defined as the distance between the centroids of the top and bottom connecting surfaces. One can calculate them by doubling the averages of the `z' coordinates calculated in Section \ref{sec:Kinematics}. Therefore, the slant height for 1 pop-out is given by:}

\begin{equation}
    \textcolor{red}{    (d_\text{gold})_{1-p/o} = 2 \left(\frac{A_z + B_z + C_z}{3}\right) = 2 \left(\frac{w + 2w\sin\theta}{3}\right) = \frac{1}{3\varphi} \approx 0.206.    }
\end{equation}

\textcolor{red}{Similarly, for 2 pop-out:}

\begin{equation}
    \textcolor{red}{(d_\text{gold})_{2-p/o} = 2 \left(\frac{A_z + B_z + C_z}{3}\right) = 2 \left(\frac{w\sin\theta + 2w}{3}\right) = \frac{2}{3\varphi} \approx 0.412.}   
\end{equation}

\textcolor{red}{As the 3 pop-out configuration (111) is fully opened, the total height is $2w$, more specifically,} 
\begin{equation}
    \textcolor{red}{(d_\text{gold})_{3-p/o} = \tan(\beta) = \frac{1}{\varphi} \approx 0.618}    
\end{equation}

\ack{V. Deshpande., Y. Phalak. Z. Zhou. and S. Li acknowledge the gracious support of the National Science Foundation (CMMI-2240211 and 2312422) and Virginia Tech (via startup funds and graduate assistantship). I. Walker acknowledges the gracious support of the National Science Foundation (CMMI-2312423 and HCC-2221126).}

\conflict{The authors declare no conflict of interests.}

\printbibliography

@article{fenci2017,
  title={Deployable structures classification: A review},
  author={Fenci, Giulia E and Currie, Neil GR},
  journal={International journal of space structures},
  volume={32},
  number={2},
  pages={112--130},
  year={2017},
  publisher={SAGE Publications Sage UK: London, England}
}

@article{puig2010,
  title={A review on large deployable structures for astrophysics missions},
  author={Puig, L and Barton, A and Rando, N},
  journal={Acta astronautica},
  volume={67},
  number={1-2},
  pages={12--26},
  year={2010},
  publisher={Elsevier}
}

@article{friedman2011a,
  title={Investigation of highly flexible, deployable structures: review, modelling, control, experiments and application},
  author={Friedman, No{\'e}mi},
  year={2011},
  publisher={{\'E}cole normale sup{\'e}rieure de Cachan-ENS Cachan; Budapesti m{\H{u}}szaki {\'e}s~…}
}

@article{santiago2013,
  title={Advances in deployable structures and surfaces for large apertures in space},
  author={Santiago-Prowald, J and Baier, H},
  journal={CEAS Space Journal},
  volume={5},
  pages={89--115},
  year={2013},
  publisher={Springer}
}

@article{zhang2021,
  title={Deployable structures: structural design and static/dynamic analysis},
  author={Zhang, Xiao and Nie, Rui and Chen, Yan and He, Baiyan},
  journal={Journal of Elasticity},
  pages={1--37},
  year={2021},
  publisher={Springer}
}

@article{liu2022a,
  title={1U-sized deployable space manipulator for future on-orbit servicing, assembly, and manufacturing},
  author={Liu, Jinguo and Zhao, Pengyuan and Chen, Keli and Zhang, Xin and Zhang, Xiang},
  journal={Space: Science \& Technology},
  year={2022},
  publisher={AAAS}
}

@article{yang2023,
  title={Volume optimisation of multi-stable origami bellows for deployable space habitats},
  author={Yang, Mengzhu and Defillion, Joe and Scarpa, Fabrizio and Schenk, Mark},
  journal={Acta Mechanica Solida Sinica},
  volume={36},
  number={4},
  pages={514--530},
  year={2023},
  publisher={Springer}
}

@article{liu2022b,
  title={Folding behaviour of a deployable composite cabin for space habitats-part 1: Experimental and numerical investigation},
  author={Liu, Tian-Wei and Bai, Jiang-Bo},
  journal={Composite Structures},
  volume={302},
  pages={116244},
  year={2022},
  publisher={Elsevier}
}

@inproceedings{hill2010,
  title={Deployment of inflatable space habitat models},
  author={Hill, Jeremy and Jacob, Jamey},
  booktitle={48th AIAA Aerospace Sciences Meeting Including the New Horizons Forum and Aerospace Exposition},
  pages={793},
  year={2010}
}

@inproceedings{ekblaw2019,
  title={Space habitat reconfigurability: TESSERAE platform for self-aware assembly},
  author={Ekblaw, Ariel and Prosina, Anastasia and Newman, Dava and Paradiso, Joseph},
  booktitle={Proc. 30th IAA Symp. Space Soc},
  year={2019}
}

@article{thrall2014,
  title={Accordion shelters: A historical review of origami-like deployable shelters developed by the US military},
  author={Thrall, Ashley P and Quaglia, CP},
  journal={Engineering structures},
  volume={59},
  pages={686--692},
  year={2014},
  publisher={Elsevier}
}

@article{mira2014,
  title={Deployable scissor arch for transitional shelters},
  author={Mira, Lara Alegria and Thrall, Ashley P and De Temmerman, Niels},
  journal={Automation in Construction},
  volume={43},
  pages={123--131},
  year={2014},
  publisher={Elsevier}
}

@article{troise2021,
  title={Preliminary analysis of a lightweight and deployable soft robot for space applications},
  author={Troise, Mario and Gaidano, Matteo and Palmieri, Pierpaolo and Mauro, Stefano},
  journal={Applied Sciences},
  volume={11},
  number={6},
  pages={2558},
  year={2021},
  publisher={MDPI}
}

@article{sedal2020,
  title={Design of deployable soft robots through plastic deformation of kirigami structures},
  author={Sedal, Audrey and Memar, Amirhossein H and Liu, Tianshu and Meng{\"u}{\c{c}}, Yi{\u{g}}it and Corson, Nick},
  journal={IEEE Robotics and Automation Letters},
  volume={5},
  number={2},
  pages={2272--2279},
  year={2020},
  publisher={IEEE}
}

@article{blumenschein2018,
  title={A tip-extending soft robot enables reconfigurable and deployable antennas},
  author={Blumenschein, Laura H and Gan, Lucia T and Fan, Jonathan A and Okamura, Allison M and Hawkes, Elliot W},
  journal={IEEE Robotics and Automation Letters},
  volume={3},
  number={2},
  pages={949--956},
  year={2018},
  publisher={IEEE}
}

@article{friedman2011b,
  title={Investigation of highly flexible, deployable structures: review, modelling, control, experiments and application},
  author={Friedman, No{\'e}mi},
  year={2011},
  publisher={{\'E}cole normale sup{\'e}rieure de Cachan-ENS Cachan; Budapesti m{\H{u}}szaki {\'e}s~…}
}

@article{zhang2020,
  title={A pneumatic/cable-driven hybrid linear actuator with combined structure of origami chambers and deployable mechanism},
  author={Zhang, Zhuang and Chen, Genliang and Wu, Haiyu and Kong, Lingyu and Wang, Hao},
  journal={IEEE Robotics and Automation Letters},
  volume={5},
  number={2},
  pages={3564--3571},
  year={2020},
  publisher={IEEE}
}

@inproceedings{schenk2013,
  title={Inflatable cylinders for deployable space structures},
  author={Schenk, Mark and Kerr, SG and Smyth, AM and Guest, Simon D},
  booktitle={First Conference Transformables},
  volume={9},
  pages={1--6},
  year={2013},
  organization={Seville Spain, Sept}
}

@article{li2015b,
  title={Fluidic origami with embedded pressure dependent multi-stability: a plant inspired innovation},
  author={Li, Suyi and Wang, KW},
  journal={Journal of The Royal Society Interface},
  volume={12},
  number={111},
  pages={20150639},
  year={2015},
  publisher={The Royal Society}
}

@inproceedings{zirbel2014,
  title={Deployment methods for an origami-inspired rigid-foldable array},
  author={Zirbel, Shannon A and Trease, Brian P and Magleby, Spencer P and Howell, Larry L},
  booktitle={The 42nd Aerospace Mechanism Symposium},
  year={2014}
}

@article{filipov2015,
  title={Origami tubes assembled into stiff, yet reconfigurable structures and metamaterials},
  author={Filipov, Evgueni T and Tachi, Tomohiro and Paulino, Glaucio H},
  journal={Proceedings of the National Academy of Sciences},
  volume={112},
  number={40},
  pages={12321--12326},
  year={2015},
  publisher={National Acad Sciences}
}

@article{chen2015,
  title={Origami of thick panels},
  author={Chen, Yan and Peng, Rui and You, Zhong},
  journal={Science},
  volume={349},
  number={6246},
  pages={396--400},
  year={2015},
  publisher={American Association for the Advancement of Science}
}

@article{melancon2021,
  title={Multistable inflatable origami structures at the metre scale},
  author={Melancon, David and Gorissen, Benjamin and Garc{\'\i}a-Mora, Carlos J and Hoberman, Chuck and Bertoldi, Katia},
  journal={Nature},
  volume={592},
  number={7855},
  pages={545--550},
  year={2021},
  publisher={Nature Publishing Group UK London}
}

@inproceedings{fernandez2017,
  title={Advanced deployable shell-based composite booms for small satellite structural applications including solar sails},
  author={Fernandez, Juan M},
  booktitle={International Symposium on Solar Sailing 2017},
  number={NF1676L-25486},
  year={2017}
}

@article{sickinger2006,
  title={Structural engineering on deployable CFRP booms for a solar propelled sailcraft},
  author={Sickinger, Christoph and Herbeck, Lars and Breitbach, Elmar},
  journal={Acta Astronautica},
  volume={58},
  number={4},
  pages={185--196},
  year={2006},
  publisher={Elsevier}
}

@article{liu2017,
  title={Programmable deployment of tensegrity structures by stimulus-responsive polymers},
  author={Liu, Ke and Wu, Jiangtao and Paulino, Glaucio H and Qi, H Jerry},
  journal={Scientific reports},
  volume={7},
  number={1},
  pages={3511},
  year={2017},
  publisher={Nature Publishing Group UK London}
}

@article{kaufmann2022,
  title={Harnessing the multistability of kresling origami for reconfigurable articulation in soft robotic arms},
  author={Kaufmann, Joshua and Bhovad, Priyanka and Li, Suyi},
  journal={Soft Robotics},
  volume={9},
  number={2},
  pages={212--223},
  year={2022},
  publisher={Mary Ann Liebert, Inc., publishers 140 Huguenot Street, 3rd Floor New~…}
}

@article{wang2017,
  title={Kirigami/origami-based soft deployable reflector for optical beam steering},
  author={Wang, Wei and Li, Chenzhe and Rodrigue, Hugo and Yuan, Fengpei and Han, Min-Woo and Cho, Maenghyo and Ahn, Sung-Hoon},
  journal={Advanced Functional Materials},
  volume={27},
  number={7},
  pages={1604214},
  year={2017},
  publisher={Wiley Online Library}
}

@article{sun2022,
  title={A photoorganizable triple shape memory polymer for deployable devices},
  author={Sun, Jiahao and Peng, Bo and Lu, Yao and Zhang, Xiao and Wei, Jia and Zhu, Chongyu and Yu, Yanlei},
  journal={Small},
  volume={18},
  number={9},
  pages={2106443},
  year={2022},
  publisher={Wiley Online Library}
}

@article{meng2022,
  title={Deployable mechanical metamaterials with multistep programmable transformation},
  author={Meng, Zhiqiang and Liu, Mingchao and Yan, Hujie and Genin, Guy M and Chen, Chang Qing},
  journal={Science Advances},
  volume={8},
  number={23},
  pages={eabn5460},
  year={2022},
  publisher={American Association for the Advancement of Science}
}

@article{bichara2023,
  title={A multi-stable deployable quadrifilar helix antenna with radiation reconfigurability for disaster-prone areas},
  author={Bichara, Rosette Maria and Costantine, Joseph and Tawk, Youssef and Sakovsky, Maria},
  journal={Nature Communications},
  volume={14},
  number={1},
  pages={8511},
  year={2023},
  publisher={Nature Publishing Group UK London}
}

@article{haghpanah2016,
  title={Multistable shape-reconfigurable architected materials},
  author={Haghpanah, Babak and Salari-Sharif, Ladan and Pourrajab, Peyman and Hopkins, Jonathan and Valdevit, Lorenzo},
  journal={Adv. Mater},
  volume={28},
  number={36},
  pages={7915--7920},
  year={2016}
}

@article{li2015a,
  title={Fluidic origami: a plant-inspired adaptive structure with shape morphing and stiffness tuning},
  author={Li, Suyi and Wang, KW},
  journal={Smart Materials and Structures},
  volume={24},
  number={10},
  pages={105031},
  year={2015},
  publisher={IOP Publishing}
}

@article{zhai2018,
  title={Origami-inspired, on-demand deployable and collapsible mechanical metamaterials with tunable stiffness},
  author={Zhai, Zirui and Wang, Yong and Jiang, Hanqing},
  journal={Proceedings of the National Academy of Sciences},
  volume={115},
  number={9},
  pages={2032--2037},
  year={2018},
  publisher={National Acad Sciences}
}

@inproceedings{daye2023,
  title={Active deployment of ultra-thin composite booms with piezoelectric actuation},
  author={Daye, Jacob G and Lee, Andrew J},
  booktitle={Active and Passive Smart Structures and Integrated Systems XVII},
  volume={12483},
  pages={94--103},
  year={2023},
  organization={SPIE}
}

@article{zhang2021b,
  title={Yoshimura-origami based earthworm-like robot with 3-dimensional locomotion capability},
  author={Zhang, Qiwei and Fang, Hongbin and Xu, Jian},
  journal={Frontiers in Robotics and AI},
  volume={8},
  pages={738214},
  year={2021},
  publisher={Frontiers Media SA}
}

@article{cai2016,
  title={Motion analysis of a foldable barrel vault based on regular and irregular Yoshimura origami},
  author={Cai, Jianguo and Deng, Xiaowei and Xu, Yixiang and Feng, Jian},
  journal={Journal of Mechanisms and Robotics},
  volume={8},
  number={2},
  pages={021017},
  year={2016},
  publisher={American Society of Mechanical Engineers}
}

@article{jiang2022,
  title={Parametric design of developable structure based on yoshimura origami pattern},
  author={Jiang, Haolei and Liu, W and Huang, Haoyu and Wang, Yiqin},
  journal={Sustainable Structures},
  year={2022},
  publisher={Newcastle University}
}

@article{pratapa2022,
  title={Thick panel origami for load-bearing deployable structures},
  author={Pratapa, Phanisri P and Bellamkonda, Abhilash},
  journal={Mechanics Research Communications},
  volume={124},
  pages={103937},
  year={2022},
  publisher={Elsevier}
}

@article{seo2021,
  title={Origami-structured actuating modules for upper limb support},
  author={Seo, Seongmin and Park, Wookeun and Lee, Dongman and Bae, Joonbum},
  journal={IEEE Robotics and Automation Letters},
  volume={6},
  number={3},
  pages={5239--5246},
  year={2021},
  publisher={IEEE}
}

@article{gimenez2024,
  title={Shear and shear-induced normal responses of origami cylinders relate to their structural asymmetries},
  author={Gim{\'e}nez-Ribes, Gerard and Ghorbani, Aref and Teng, Soon Yuan and van der Linden, Erik and Habibi, Mehdi},
  journal={Materials \& Design},
  volume={240},
  pages={112874},
  year={2024},
  publisher={Elsevier}
}

@article{seffen2014,
  title={Surface texturing through cylinder buckling},
  author={Seffen, KA and Stott, SV},
  journal={Journal of Applied Mechanics},
  volume={81},
  number={6},
  pages={061001},
  year={2014},
  publisher={American Society of Mechanical Engineers}
}

@article{xue2024rigid,
  title={Rigid-flexible coupled origami robots via multimaterial 3D printing},
  author={Xue, Wenbo and Sun, Zechu and Ye, Haitao and Liu, Qingjiang and Jian, Bingcong and Wang, Yanjie and Fang, Hongbing and Ge, Qi},
  journal={Smart Materials and Structures},
  volume={33},
  number={3},
  pages={035004},
  year={2024},
  publisher={IOP Publishing}
}

@article{yang2024novel,
  title={A Novel Data Augmentation Method Based on Denoising Diffusion Probabilistic Model for Fault Diagnosis Under Imbalanced Data},
  author={Yang, Xiongyan and Ye, Tianyi and Yuan, Xianfeng and Zhu, Weijie and Mei, Xiaoxue and Zhou, Fengyu},
  journal={IEEE Transactions on Industrial Informatics},
  year={2024},
  publisher={IEEE}
}

@book{craig2006introduction,
  title={Introduction to robotics},
  author={Craig, John J},
  year={2006},
  publisher={Pearson Educacion}
}

@inproceedings{xu2021design,
  title={Design of lightweight and extensible tendon-driven continuum robots using origami patterns},
  author={Xu, Yunti and Peyron, Quentin and Kim, Jongwoo and Burgner-Kahrs, Jessica},
  booktitle={2021 IEEE 4th International Conference on Soft Robotics (RoboSoft)},
  pages={308--314},
  year={2021},
  organization={IEEE}
}

@article{zhang2016extensible,
  title={An extensible continuum robot with integrated origami parallel modules},
  author={Zhang, Ketao and Qiu, Chen and Dai, Jian S},
  journal={Journal of Mechanisms and Robotics},
  volume={8},
  number={3},
  pages={031010},
  year={2016},
  publisher={American Society of Mechanical Engineers}
}

@inproceedings{singer1982status,
  title={The status of experimental buckling investigations of shells},
  author={Singer, J},
  booktitle={Buckling of Shells: Proceedings of a State-of-the-Art Colloquium, Universit{\"a}t Stuttgart, Germany, May 6--7, 1982},
  pages={501--533},
  year={1982},
  organization={Springer}
}

\end{document}